\documentclass[12pt]{article}

\usepackage{latexsym}
\long\def\nop#1{}

\def\comment{\edef\cps{\the\parskip} \parskip=0.5cm \begingroup \tt}

\expandafter\ifx\csname shortcite\endcsname\relax
\let\shortcite=\cite
\fi

\newbox\current

\long\def\plframebox#1{
\setbox\current\vbox{#1}		% set box

\vbox to \ht\current {\hrule\vss
\hbox to \wd\current {%
\vrule \hss\box\current\hss \vrule}
\vss\hrule }
}

\long\def\eatpar#1{%
\ifx#1\par                      % se il token e' \par
\let\nextmove=\eatpar           % rimetti \eatpar in coda
\else
\let\nextmove=#1%               altrimenti, rimetti il token in coda
\fi
\noexpand\nextmove%             il token o \eatpar viene rimesso in coda
}

\def\modifymargins#1#2{
\newdimen\addtoh
\newdimen\addtow
\addtoh=#1
\addtow=#2

\advance\topmargin by -\addtoh
\multiply\addtoh by 2
\advance\textheight by \addtoh

\advance\oddsidemargin by -\addtow
\advance\evensidemargin by -\addtow
\multiply\addtow by 2
\advance\textwidth by \addtow
}

\begingroup
\catcode`\~=11
\gdef\centertilde#1{\lower #1pt\hbox{~}}
\endgroup

\newcount\currenttime
\newcount\hour
\newcount\minute

\def\printtime{%
\currenttime=\time
\hour=\currenttime
\divide\hour by 60
\minute=-\hour
\multiply\minute by 60
\advance\minute by \currenttime
\the\hour:\ifnum\minute<10 0\fi\the\minute
}

\begingroup
\makeatletter
\global\let\@@date=\@date
\gdef\@date{\@@date\ --- \printtime}
\endgroup

\def\oggi{\number\day\space 
\ifcase\month\or
Gennaio\or Febbraio\or Marzo\or Aprile\or Maggio\or Giugno\or
Luglio\or Agosto\or Settembre\or Ottobre\or Novembre\or Dicembre\fi
\space \number\year}

\newcounter{rmexample}

\def\proof{\noindent {\sl Proof.\ \ }}

\def\qed{\hfill{\boxit{}}
  \ifdim\lastskip<\medskipamount \removelastskip\penalty55\medskip\fi}
\def\qedn#1{\hfill{\boxit{}$_#1$}
  \ifdim\lastskip<\medskipamount \removelastskip\penalty55\medskip\fi}
\long\def\boxit#1{\vbox{\hrule\hbox{\vrule\kern3pt
                  \vbox{\kern3pt#1\kern3pt}\kern3pt\vrule}\hrule}}

  \def\D{{\cal D}}

\def\ie{i.e.}
\def\eg{e.g.}

\def\imply{\rightarrow}

\def\l{\langle}
\def\r{\rangle}
\newcommand{\pair}[2]{\langle #1, #2 \rangle}

\def\var{V\!ar}

\def\true{{\sf true}}

\def\C{{\rm C}}

\def\p{{\rm P}}
\def\np{{\rm NP}}
\def\conp{{\rm coNP}}
\def\Dp{${\rm D}^p$}
\def\S#1{\mbox{$\Sigma^p_{#1}$}}

\def\D#1{\mbox{$\Delta^p_{#1}$}}
\def\Dlog#1{\mbox{$\Delta^p_{#1}[\log n]$}}

\def\poly{{\rm poly}}

\def\nuc#1{\mbox{$\parallel\!\leadsto$#1}}

\def\nucC{\nuc{\rm C}}
\def\nucp{\nuc{\rm P}}
\def\nucnp{\nuc{\rm NP}}
\def\nucconp{\nuc{\rm coNP}}
\def\nucS#1{\nuc{$\Sigma^p_{#1}$}}
\def\nucP#1{\nuc{$\Pi^p_{#1}$}}
\def\nucD#1{\nuc{\D{#1}}}
\def\nucDp{\nuc\Dp}
\def\nucDlog#1{{\rm nucomp}\Dlog{#1}}

\def\nucomp{\leq_{nucomp}}

\def\profont{\sf}

\def\x3c{{\profont x3c}}

\def\possnewtheorem#1#2{
\expandafter\ifx\csname #1\endcsname\relax
\newtheorem{#1}{#2}
\fi
}

\def\possnewtheoremthree#1[#2]#3{
\expandafter\ifx\csname #1\endcsname\relax
\newtheorem{#1}[#2]{#3}
\fi
}

\possnewtheorem{theorem}{Theorem}
\possnewtheorem{corollary}{Corollary}
\possnewtheorem{lemma}{Lemma}
\possnewtheoremthree{proposition}[theorem]{Proposition}
\possnewtheorem{definition}{Definition}
\possnewtheorem{question}{Question}
\possnewtheorem{example}{Example}
\possnewtheorem{property}{Property}
\possnewtheorem{assumption}{Assumption}
\possnewtheorem{conjecture}{Conjecture}
\newenvironment{theorem*}[1]{{\noindent \bf Theorem~#1}\begin{it}}{\end{it}\

}

\let\shortcite=\cite

\title{Compilability of Abduction}
\author{Paolo Liberatore\\
\normalsize Dipartimento di Informatica e Sistemistica\\
\normalsize Universit\`a di Roma ``La Sapienza''\\
\normalsize Via Salaria 113, 00198 Roma - Italy\\
\normalsize Email: {\tt paolo@liberatore.org}
\and Marco Schaerf\\
\normalsize Dipartimento di Informatica e Sistemistica\\
\normalsize Universit\`a di Roma ``La Sapienza''\\
\normalsize Via Salaria 113, 00198 Roma - Italy\\
\normalsize Email: {\tt schaerf@dis.uniroma1.it}
}

\date{}

\begin{document}

\maketitle

%%%%%%%%%%%%% abst %%%%%%%%%%%%%%
\begin{abstract}

Abduction is one of the most important forms of reasoning; it has
been successfully applied to several practical problems such as
diagnosis. In this paper we investigate whether the computational
complexity of abduction can be reduced by an appropriate use of
preprocessing. This is motivated by the fact that
part of the data of the problem (namely, the set of all possible
assumptions and the theory relating assumptions and manifestations)
are often known before the rest of the problem. In this paper,
we show some complexity results about abduction when compilation
is allowed.

\end{abstract}

\newpage

\tableofcontents

\let\sectionnewpage=\newleaf
\let\subsectionnewpage=\relax

%%%%%%%%%%%%% intr %%%%%%%%%%%%%%
\section{Introduction}

Deduction, induction, and abduction~\shortcite{peir-55} are the
three basic reasoning mechanisms. Deduction allows drawing
conclusions from known facts using some piece of knowledge, so
that ``battery is down'' allows concluding ``car will not start''
thanks to the knowledge of the rule ``if the battery is down,
the car will not start''. Induction derives rules from the
facts: from the fact that the battery is down and that the
car is not starting up, we may conclude the rule relating these
two facts. Abduction is the inverse of deduction
(to some extent \cite{cial-96}): from the
fact that the car is not starting up, we conclude that the
battery is down. Clearly, this is not the only possible
explanation of a car not starting up. Therefore, we may get
more than one explanation. This is an important difference
between abduction and deduction, making the former, in
general, more complex.

While deduction formalizes the process of drawing conclusions,
abduction formalizes the diagnostic process, which attempts to
invert the cause-effect relation by inferring the causes from
its observable effects. The example of the car shows such
an application: complete knowledge about car would allow
finding (\ie, abducing) the possible reasons of why the car
is not starting up. The following example shows how abduction
can be applied to formalize a diagnostic scenario.

\begin{example}

While writing a paper with some authors located in another
country, you get a set of macros that are used in a nice
figure they drew. However, when compiling the .tex file, an
incomprehensible error message results. Four explanations
are possible:

\begin{description}

\item[$a$]: the macro has been used with the wrong arguments;

\item[$p$]: the package $X$ is required;

\item[$t$]: the macro is incompatible with package $X$;

\item[$v$]: the wrong version of TeX has been used.

\end{description}

This scenario can be formalized in logical terms by 
introducing a variable $f$ to denote the presence of
compile errors: since each of the facts above causes
$f$, we know $a \rightarrow f$, $p \rightarrow f$, etc.
Moreover, we know that a package cannot at the same
time be required and incompatible with the macros.
The following theory $T$ formalize our knowledge.

\[
T = \{ a \imply f, p \imply f, t \imply f,
v \imply f, \neg (p \wedge t) \}
\]

This theory relates the observed effect (the compile
error) with its possible causes (we used the wrong
version of TeX, etc.) Therefore, it can be used to
find the possible causes: namely, an explanation is
a set of facts that logically imply the observed
effect. Formally, an explanation is a set of variable
that allow deriving the observed effects from the
theory $T$. However, to make sense an explanation
has to be consistent with our knowledge, that is,
with the theory $T$.

\end{example}

This example shows that a given problem of abduction may
have one, none, or even many possible solutions (explanations).
Moreover, a consistent and an implication checks are required
just to verify an explanation. These facts intuitively explain
why abduction is to be expected to be harder than deduction.
This observation has indeed been confirmed by theoretical
results. Selman and Levesque~\cite{selm-leve-90}
and Bylander {\em et al.}~\shortcite{byla-etal-89}
proved the first results about fragments of abductive reasoning,
Eiter and Gottlob~\shortcite{eite-gott-95-a} presented
an extensive analysis, and Eiter and Makino have shown
the complexity of computing all abductive explanations
\cite{eite-maki-02}. All these results proved that abduction
is, in general, harder than deduction. The analysis has also shown
that several problems are of interest in abduction. Not only the
problem of finding an explanation is relevant, but also the
problems of checking an explanation, or whether a fact is
in all, or some, of the explanations are.

A common fact about deduction and abduction is that the
knowledge relating facts may be known in advance, while
the particular observation may change from time to time.
In the example of the car, the fact that the dead battery
makes the car not to start is always known, while the
fact that the battery is dead may or may not be true. The
possible causes of TeX errors are known before a specific
error message comes out, etc.

We can therefore assign two different statuses to the knowledge
base and to the single facts: while the knowledge base is
{\em fixed}, the single facts are {\em varying}. In the
example above, $T$ will always reflect the state of the
word, while $f$ is only true when the TeX complains
about something.

This difference has computational consequences. While
the example we have shown here does not present any problem
of efficiency, larger and more complex abduction
problems result from the formalization of real-world domains.
The difference of status of $T$ and the observations
can then be exploited. Indeed, since $T$ is always the
same, we can perform a preprocessing step on it alone,
even before the status of the observations are known.
Clearly, we cannot explain an observation we do not
know. However, this preprocessing step can be used to
perform some computation that would otherwise be done
on $T$ alone. As a result, finding a solution might
take less time when the observation finally get known.

The idea of using a preprocessing step for speeding-up
the solving of abduction problems is not new. For instance,
Console, Portinale, and Dupr\`e~\shortcite{cons-port-dupr-96}
have shown how compiled knowledge can be used in the process
of abductive diagnosis.

Preprocessing part of the input data has also been used in
many other areas of computer science, as there are many
problems with a similar fixed-varying part pattern. However,
the first formalization of intractability with preprocessing
is relatively recent \cite{cado-etal-02}. In this paper,
we characterize the complexity of the problems about
abductions from this point of view.

%%%%%%%%%%%%% prel %%%%%%%%%%%%%%
\section{Preliminaries}

The problem of abduction is formalized by a knowledge
base, a set of observations, and a set of possible
facts that can explain the observations. In this paper,
we are only concerned about propositional logic.
Therefore, the knowledge is formalized by a propositional
theory $T$. We usually denote by $M$ the set of
observations.

The theory is $T$ must necessarily contain all variables
of $M$, otherwise there would be no way of explaining
the observations. In general, the theory $T$ contains other
variables as well, describing facts we do not know whether
they are true or not. Some of these facts can be taken
as part of a possible explanation, while others are can
not. Intuitively, when we are trying to establish the
causes of an observation, we want the {\em first cause},
and not something that is only a consequence of it.
In the example of the car, the fact that there is no
voltage in the starting engine explains the fact that
the car is not starting up, but it is not an acceptable
explanation, as it does not tell where the real problem
is (the battery). Therefore, the abduction problem is
not defined only in terms of the theory and the
observation, but also of the set of possible facts
(variable) we would accept as first causes of the
observation.

Formally, an instance of abduction is a triple
$\l H, M, T \r$. The observations are formalized as $M$,
which is a set of variables. $T$ is a propositional
theory formalizing our knowledge of the domain. Finally,
$H$ is a set of variables; these variables are the ones
formalizing facts that we regards as possible first
causes.

Abduction is the process of explaining the observation.
Its outcome will therefore be a set of facts from which
all observations can be inferred. Since we can only 
use variables of $H$ to form explanations, these will
be subsets $H' \subseteq H$. Moreover, an explanation
can only be accepted if it is consistent with our
knowledge. This leads to the following definition of
the possible solutions (explanations) of a given
abduction problem  $\l H, M, T \r$.

\[
SOL( H, M, T ) =
\{ H' \subseteq H  ~|~	H' \cup T \mbox{ is consistent and }
			H' \cup T \models M \}
\]

We apply this definition to the running example of the
TeX file.

\begin{example}

The propositional theory of the example shown in the
introduction is $T = \{ a \imply f, p \imply f, t \imply f,
v \imply f, \neg (p \wedge t) \}$. The observation is
the variable formalizing the presence of compiler errors,
that is, $M=\{f\}$. Of the variables of $T$, all but
$T$ can be taken as possible first causes of the problem,
that is, $H=\{a, p, t, v\}$.

Abduction amounts to finding a set of literals that
explain the observation $f$. Formally, this is captured
by the constraint $H' \cup T \models M$. Note that
$H' = \{f\}$ satisfies this formula; this is not
an acceptable explanation: ``the reason of why the file
does not compile is that it does not compile'' is a
tautology, not an explanation. This problem is avoided
by enforcing $H' \subseteq H$.

All non-empty subsets of $H$ implies, together with $T$,
the observation $M$. However, the subsets containing
both $p$ and $t$ are inconsistent with $T$. Therefore,
the set of solution of the problem is given by:

\[
SOL( H, M, T ) =
\{ H' \subseteq H ~|~ H \not= \emptyset ,~ \{t,p\} \not\subseteq H' \}
\]

This is simply the formal result of our current definition.
However, some explanations in this set are not really reasonable:
for example, the explanation is $\{a, t, v\}$ seems overly pessimistic:
the macro has been called in the wrong way {\em and} a package
is required {\em and} we used the wrong compiler version.

\end{example}

The set $SOL( H, M, T )$ contains all explanations we
consider possible. However, some explanations may be
more likely than others. For example, explanations
requiring a large number of assumptions are often
less likely than explanations with less assumptions.

Likeliness of explanations is formalized by an
an ordering $\preceq$ over the subsets of $H$.
Given a specific $\preceq$, the set of minimal
solutions is defined as follows.\eatpar

\[
SOL_\preceq( H, M, T ) = \min( SOL(H, M, T), \preceq )
\]

The ordering $\preceq$ is used to formalize the
relative plausibility explanations: $H' \prec H''$
means that $H'$ is considered more likely to
be the ``real'' cause of the manifestations
than $H''$. The ordering $\preceq$ represents the
concept of ``at least as likely as'', thus
$H' \cong H''$ holds if $H'$ and $H''$ are equally
likely. The definition of $SOL_\preceq$ formalizes
the principle of choosing only the explanations we
consider more likely.

An implicit assumption of this definitions is that
the ordering $\preceq$ does not depend on the set of 
manifestations. We also assume that $\preceq$ is a
``well-founded'' ordering, that is, any non-empty
set of explanations has at least one $\preceq$-minimal
element. Therefore, if the set $SOL( H, M, T )$ is
not empty, then $\min(SOL( H, M, T ),\preceq)$ is not
empty as well.

In this paper we take into account several plausibility
ordering. The absence of a preference among the explanations
can be formalized as the ordering $\preceq$ that is equal
to the universal relation, that is, $H' \preceq H''$ for
any pair of sets of variables $H'$ and $H''$.

Besides this no-information ordering, the two simplest
and most natural orderings are $\subseteq$-preference,
where an explanation $H_1$ is more likely of $H_2$ if
$H_1 \subseteq H_2$, and $\leq$-preference, where $H_1$
is preferred to $H_2$ if it contains less hypothesis,
that is, $|H_1| \leq |H_2|$.

Both these orderings are based on the principle of making
as few hypotheses as possible, and by assuming that all
hypotheses are equally likely. Two other orderings follows
from assuming that the hypotheses are not equally likely:
the $\subseteq$-prioritization and the $\leq$-prioritization.

In particular, we assume that the hypotheses are partitioned
into equivalence classes of equal likeliness.
Let $\l H_1,\ldots,H_m \r$ be such a partition. By definition,
it holds $H_1 \cup \cdots \cup H_m = H$ and
$H_i \cap H_j = \emptyset$ for each $i
\not= j$. The instances of the problem of abduction can thus be
written as $\l \l H_1,\ldots,H_m \r, M, T \r$.  The set of all
assumptions $H$ is implicitly defined as the union of the classes
$H_i$. We assume
that the hypotheses in $H_1$ are the most likely, while those
in $H_m$ are the least likely.

The $\subseteq$-prioritization and $\leq$-prioritization compare
explanations on the basis of their relative plausibility.
Namely, the explanations that use hypothesis in lower classes
are more likely than explanations using hypothesis in higher classes.
This idea, when combined with subset containment, defines the
$\subseteq$-prioritization. When it is combined with the
cardinality-based ordering, it defines the $\leq$-prioritization.
Formal definition is below.

Penalization is the last form of preference we consider. The
idea is to assign weights to assumptions to formalize their
likeliness. Explanations with the least total weight are
preferred. Weights encodes the likeliness of assumptions:
the most high the weight of an assumption, the
unlikely it is to be true. To use penalization, the instance
of the problem must include, besides $H$, $M$, and $T$,
an $n$-tuple of weights $W=\l w_1,\ldots,w_n \r$, where each
$w_i$ is an integer number (the weight) associated to a
variable $h_i \in H$.  The instance can thus be written
$\l W, H, M, T \r$.

The considered orderings are formally defined as follows:

\begin{description}

\item[$\subseteq$-preference] $H' \preceq H''$ if and only if
$H' \subseteq H''$;

\item[$\leq$-preference] $H' \preceq H''$ if and only if $|H'| \leq |H''|$;

\item[$\subseteq$-prioritization] $H' \preceq H''$ if and only if $H'=H''$
or there exists $i$ such that $H' \cap H_m = H'' \cap H_m$, $\ldots$, 
$H' \cap H_i = H'' \cap H_i$, $H' \cap H_{i-1} \subset H'' \cap H_{i-1}$;

\item[$\leq$-prioritization] $H' \preceq H''$ if and only if either $|H' \cap
H_i| =|H'' \cap H_i|$ for each $i$, or there exists $i$ such that $|H' \cap
H_m| = |H'' \cap H_m|$, $\ldots$, $|H' \cap H_i| = |H'' \cap H_i|$, $|H' \cap
H_{i-1}| \leq |H'' \cap H_{i-1}|$;

\item[penalization] $H' \preceq H''$ if and only if
$\sum_{h_i \in H'} w_i \leq \sum_{h_j \in H''} w_j$.

\end{description}

Let us consider the use of these orderings on
the running example.

\begin{example}

The use of $\subseteq$-preference or $\leq$-preference
reduces the set of possible explanations of the example
of the TeX file. Namely, $\leq$-preference let 
minimal-size explanations only to be solutions of the
problem. The only such explanations are
$\{a\}$, $\{p\}$, $\{t\}$, and $\{v\}$. The explanation
$\{a, t, v\}$, being not minimal, is not a solution
of the problem any more. The use of preference therefore
avoids having as solutions some sets that contains too
many hypotheses. Since $\subseteq$-preference only
selects explanations that are not contained in other
ones, the only solutions it produces are 
$\{a\}$, $\{p\}$, $\{t\}$, and $\{v\}$. In this case,
the two kinds of the preference generate the same
solutions, but this is not always the case.

Prioritization allows for a further refinement of the
set of solutions by exploiting the plausibility ordering
over the hypotheses. For example, we may assume that
the fact that package $X$ is required and that we used
the wrong version of the compiler are the two most
likely hypotheses. Formally, they will be part of the
first set of assumptions $H_1$, while the other assumptions
will therefore go in $H_2$. Formally, the problem instance
is now $\l \l H_1,H_2 \r, M, T \r$. Both
$\subseteq$-prioritization and $\leq$-prioritization
produce $\{p\}$ and $\{v\}$ as the only minimal
explanations. This is because all other explanations
either have a bigger intersection with $H_2$, or an
equal intersection with $H_2$ but a bigger intersection
with $H_1$.

Finally, penalization requires a weight (an integer
number) for each hypothesis. Let us for example use
the set of weights $\l 4, 2, 4, 1\r$ associated with
the set of hypotheses $\l a, p, t, v\}$. Since larger
weights correspond to less likely hypotheses, we are
assuming that our first and third hypotheses ($a$ and
$t$) are the least likely, while $p$ is more likely
and $v$ is the most likely. From definition, the
explanation $\{v\}$ is the one having the least weight,
and is therefore the only solution of the problem.

\end{example}

\iffalse

In the sequel, we assume that the set of all possible manifestations is
identical to the set of all variables.  In some cases, it is useful to assume
that this set is disjoint to the set of hypothesis. In such cases, a very
simple reduction can be used: we create a copy of each variable $a$, and then
we add $a \equiv a'$ to the theory. This way, we can use the original variables
in $H$ for the hypothesis and the variables $a'$ for the possible
manifestations.  Moreover, without loss of generality,

In this paper, we assume that $T$ is a 3CNF formula
(this assumption does not cause a loss of generality
unless we want to assume that $H \cup M = \var(T)$.)
We also assume that the variables of $T$ are the
assumptions $h_1,\ldots,h_m$, plus another disjoint set of variables we denote
as $x_1,\ldots,x_n$. 

\fi

The basic problem of abduction is that of finding
one or more explanations. However, we have already
remarked that none may exist. Therefore, the first
problem we consider is the existence one: given
an instance of abduction, does an explanation
exist? Another related problem is that of verifying,
once a set of hypotheses has been found, whether
it is really an explanation or not.

Other problems are related to the structure of
the explanations. Namely, hypotheses that are in
{\em all} explanations may considered as ``sure''
conclusions of the abductive process. On the other
hand, hypotheses that are part of some explanations
can be regarded as ``possible'' conclusions.

The formal definition of these questions as
decision problems is as follows.

\begin{description}
  
\item[Existence:] is there an explanation of the observed
  manifestations? That is, $SOL(H,M,T) \not= \emptyset$?

\item[Verification:] given a set $H' \subseteq H$, is $H'$ a minimal
  solution? That is, $H' \in SOL_{\preceq} (H, M, T)$?

\item[Relevance:] given a variable $h \in H$, is there a minimal solution
containing~$h$? That is, $\exists H' \subseteq H$ such that $H' \in
SOL_{\preceq} (H, M, T)$ and $h \in H'$?

\item[Necessity:] is $h \in H$ in all, and at least one, minimal
  solution?  That is, $SOL(H, M, T) \neq \emptyset$ and $\forall H'
  \subseteq H$ we have that $H' \in SOL_{\preceq} (H, M, T)$ implies
  $h \in H'$?
  
\item[Dispensability:] is $h \in H$ such that either there is no
  solution or there exists one who does not contain $h$? That is,
  $SOL(H, M, T) = \emptyset$ or $\exists H' \subseteq H$ such that $H'
  \in SOL_{\preceq} (H, M, T)$ and $h \not \in H'$?

\end{description}

Dispensability is the converse of the problem of necessity,
since an hypothesis $h$ is dispensable if and only if it
is not necessary. The problem of dispensability is not of
much interest by itself, but is sometimes useful for
simplifying the proofs.

Clearly, the ordering does not matter for the problem of existence,
since we consider only well-founded orderings: therefore,
an explanation exists if and only if a minimal explanation
exists. For the other problems, the ordering must be taken
into account. Different orderings may lead to different
computational properties.

In this paper, we assume that $T$ is a 3CNF formula:
this assumption does not cause a loss of generality
unless we want to assume that $H \cup M = \var(T)$.

%%%%%%%%%%%%% comp %%%%%%%%%%%%%%
\section{Complexity and Compilability}
\label{compilability}

The basic complexity classes of the polynomial hierarchy
\cite{stoc-76,gare-john-79}, such as \p, \np, \conp, etc.,
are assumed known to the reader. We denote by \C, $\C'$,
etc.\  arbitrary classes of the polynomial hierarchy.
The {\em length} of a string $x \in \Sigma^*$ is denoted
by $||x||$.

We summarize some definitions and results proposed to
formalize the on-line complexity of problems \cite{cado-etal-02}.
In computational complexity, problems whose solution
can only be yes or no are the most commonly analyzed. Such
problems are called {\em decision problems}. Any such
problem can be formalized as set of strings, those
whose solution is yes. For example, the problem of
propositional satisfiability (deciding whether a formula
is satisfiable or not) is characterized by the set of
the strings that represent exactly all satisfiable
formulae.

The strings that compose the set associated to a problem
represent the possible problem instances that produce a
positive solution. Problems like abduction, however,
have instances that can be naturally broken into two
parts: one part is known in advance ($T$ and $H$)
and one part is only known at run-time ($M$). Therefore,
the instances of such problems are better encoded as
{\em pairs of strings}. Therefore, a problem like abduction
is formalized by a set of pairs of strings, rather than
a set of strings. We define a {\em language of pairs} $S$
as a subset of $\Sigma^* \times \Sigma^*$. 

The difference between the first and second element of
a pair is that some preprocessing time can be spent on
the first string alone. This is done to the aim of solving
the problem faster when the second string comes to be
known. While our final aim is to reduce the running time
of this second phase, some constraints have to be put on
the preprocessing phase. Namely, we impose its result
to be of polynomial size. {\em Poly-size} function are
introduced to this purpose: a function $f$ from strings
to strings is called {\em poly-size} if there exists a
polynomial $p$ such that, for all strings $x$, it holds
$||f(x)|| \leq p(||x||)$. An exception to this definition
is when $x$ represents a natural number: in this case,
we impose $||f(x)|| \leq p(x)$. Any polynomial function is
polysize, but not viceversa. Indeed, a function $g$ is
{\em poly-time} if there exists a polynomial $q$ such that,
for all $x$, $g(x)$ can be computed in time less than or
equal to $q(||x||)$. Clearly, the running time also bounds
the size of the output string; on the other hand, even
a function requiring exponential running time can produce
a very short output. The definitions of polysize and polytime
function extend to binary functions as usual.

Using the above definitions, we introduce a new hierarchy
of classes of languages of pairs, the {\em non-uniform
compilability classes} \cite{cado-etal-02}, denoted by
\nucC, where C is a generic uniform complexity class, such
as P, \np, \conp, or \S 2.

\begin{definition}[\nucC\  classes, \cite{cado-etal-02}]

A language of pairs $S \subseteq \Sigma^* \times \Sigma^*$
belongs to \nucC\  iff there exists a binary poly-size
function $f$ and a language of pairs $S' \in \C$ such that,
for all $\pair x y \in S$, it holds:\eatpar

\[
\pair x y \in S
\mbox{~~ iff ~~}
\pair {f(x,||y||)} y \in S'
\]

\end{definition}

Clearly, any problem whose time complexity is in \C\
is also in \nucC: just take $f(x,||y||)=x$ and $S'=S$.
Some problems in \C\ however belongs to $\nucC'$ with
$\C' \subset C$; for example, some problem in \np\
are in \nucP. These are in fact the problems we are
most interested, as the preprocessing phase, running
on $x$ only, will produce $f(x)$, which allows solving
the problem in polynomial time. This is important if
these problems cannot be solved in polynomial time
without the preprocessing phase (\eg, they are
\np-complete).

The class \nucC\ generalizes the non-uniform class
\C/\poly\ --- i.e., \C/\poly\ $\subset$ \nucC\ ---
by allowing for a fixed part $x$. We extend the
definition of polynomial reduction to a concept
that can be used with these classes.

\begin{definition}[Non-uniform comp-reduction]
\label{non-uniform-reduction}

A non-uniform comp-reduction is a triple of functions
$\l f_1, f_2, g \r$, where $g$ is polytime and $f_1$
and $f_2$ are polysize. Given two problems $A$ and $B$,
$A$ is {\em non-uniformly comp-reducible\/} to $B$
(denoted by $A \nucomp B$) iff there exists a non-uniform
comp-reduction $\l f_1, f_2, g \r$ such that, for every
pair ${\pair x y }$ it holds that ${\pair x y } \in A$ if
and only if
${\pair {f_1(x,||y||)} {g(f_2(x,||y||),y)} } \in B$.

\end{definition}

These reductions allows for a concept of {\em hardness}
and {\em completeness} for the classes \nucC.

\begin{definition}[\nucC-completeness]

Let $S$ be a language of pairs and {\rm C} a complexity
class. $S$ is \nucC{\em -hard} iff for all problems
$A \in \nucC$ we have that $A \nucomp S$. Moreover,
$S$ is \nucC-complete if $S$ is in \nucC\  and
is \nucC-hard.

\end{definition}

The hierarchy formed by the compilability classes is
proper if and only if the polynomial hierarchy is proper
\cite{cado-etal-02,karp-lipt-80,yap-83} --- a fact widely
conjectured to be true.

Informally, \nucnp-hard problems are ``not compilable to P''.
Indeed, if such compilation were possible, then it would be
possible to define $f$ as the function that takes the fixed
part of the problem and gives the result of compilation
(ignoring the size of the input), and $S'$ as the language
representing the on-line processing. This would implies that a
\nucnp-hard problem is in \nucp, and this implies the collapse
of the polynomial hierarchy.  In general, a problem that is
\nucC-complete for a class \C\  can be regarded as the
``toughest'' problem in \C, in the assumption that preprocessing
the fixed part is possible.

While \nucC-completeness is adequate to show the compilability
level of a given reasoning problem, proving it requires finding
a nucomp reduction. We show a technique that let us reuse,
with simple modifications, the polytime reductions that were
used to prove the usual (uniform) hardness of the problem.
Namely, we present sufficient conditions allowing for a
polynomial reduction to imply the existence of a nucomp
reduction \cite{libe-01-jacm}.

Let us assume that we know a polynomial reduction from
the problem $A$ to the problem $B$, and we want to prove
the nucomp-hardness of $B$. Some conditions on $A$ should
hold, as well as a condition over the reduction.
If all these conditions are verified, then
there exists a nucomp reduction from $*A$ to $B$.

\begin{definition}[Classification Function]

A {\em classification function} for a problem $A$ is a
polynomial function $Class$ from instances of $A$ to
nonnegative integers, such that $Class(y) \leq ||y||$.

\end{definition}

\begin{definition}[Representative Function]

A {\rm representative function} for a problem $A$ is a
polynomial function $Repr$ from nonnegative integers to
instances of $A$, such that $Class( Repr( n) )=n$, and
that $||Repr(n)||$ is bounded by some polynomial in $n$.

\end{definition}

\begin{definition}[Extension Function]

An {\em extension function} for a problem $A$ is a polynomial function from
instances of $A$ and nonnegative integers to instances of $A$ such that, for
any $y$ and $n \geq Class(y)$, the instance $y' = Exte(y,n)$ satisfies the
following conditions:\eatpar

\begin{enumerate}
\itemsep=0pt
\item $y \in A$ if and only if $y' \in A$;
\item $Class(y')=n$.
\end{enumerate}

\end{definition}

Let us give some intuitions about these functions.
Usually, an instance of a problem is composed of a
set of objects combined in some way. For problems on
boolean formulas, we have a set of variables combined
to form a formula.  For graph problems, we have a set
of nodes, and the graph is indeed a set of edges,
which are pairs of nodes.
The classification function gives the number of objects in an
instance. The representative function thus gives an instance with the given
number of objects. This instance should be in some way ``symmetric'', in the
sense that its elements should be interchangeable (this is because the
representative function must be determined only from the number of objects.)
Possible results of the representative function can be
the set of all clauses of three literals over a given
alphabet, the complete graph over a set of nodes, the graph with no edges,
etc.

Let for example $A$ be the problem of propositional satisfiability. We can take
$Class(F)$ as the number of variables in the formula $F$, while $Repr(n)$ can
be the set of all clauses of three literals over an alphabet of $n$ variables.
Finally, a possible extension function is obtained by adding tautological
clauses to an instance.

Note that these functions are related to the problem $A$ only, and do not
involve the specific problem $B$ we want to prove hard, neither the specific
reduction used. We now define a condition over the polytime
reduction from $A$ to $B$. Since $B$ is a problem of pairs,
we can define a reduction from $A$ to $B$ as a pair of
polynomial functions $\l r,h \r$ such that $x \in A$
if and only if $\l r(x),h(x) \r \in B$.

\begin{definition}[Representative Equivalence]

Given a problem $A$ (having the above three functions), a problem of pairs
$B$, and a polynomial reduction $\langle r,h \rangle$ from $A$ to $B$, the
condition of representative equivalence holds if, for any instance $y$ of $A$,
it holds:\eatpar

\[
\l r(y),h(y) \r \in B \mbox{ ~~ iff ~~ } \l r(Repr(Class(y)),h(y) \r \in B
\]

\end{definition}

The condition of representative equivalence can be
proved to imply that the problem $B$ is \nucC-hard,
if $A$ is \C-hard \cite{libe-01-jacm}.

%%% Local Variables: 
%%% mode: latex
%%% TeX-master: "main"
%%% End: 

%%%%%%%%%%%%% noorder %%%%%%%%%%%%%%
\section{Compilability of Abduction: No Ordering}

In this section we analyze the problems of existence
of explanation, explanation verification, relevance,
and necessity, for the basic case in which no ordering
is defined. Formally, we want to determine whether
the complexity of the problems related to
$SOL(H, M, T)$ decrease thanks to the preprocessing
step on $H$ and $T$.

We first give an high-level explanation of the method
we use to prove the incompilability of the considered
problems. We begin by applying the method to the problem
of existence of explanations, and then we used it for
verification, relevance and necessity.

%%%%%%%%%%%%% method %%%%%%%%%%%%%%
\subsection{The Method}

The problem of deciding whether there exists an explanation for
a set of manifestations is \S{2}-hard \cite{eite-gott-95-a}.
Therefore, there exists a polynomial reduction from another
\S{2}-hard problem to this one. In order to prove it is also
\nucS{2}-hard we can show that the other problem has the
three functions, and the reduction satisfies the condition
of representative equivalence. Unfortunately, this is not
the case. As a result, we have to look for another reduction.

Such a reduction should be as simple as possible. In
general, the more similar two problems are, the easier
it is to find a reduction. What is the \S{2}-hard problem
that is the most similar to the problem of existence of
explanation? Clearly, the problem itself is the most similar
to itself.

The theorem of representative equivalence is indeed about
a reduction between two problems $A$ and $B$, but it does
not forbid using the same problem: it only tells that, if
we have a reduction from {\em an arbitrary} \S{2}-hard
problem $A$ to $B$, satisfying representative equivalence,
then $B$ is \nucS{2}-hard. Nothing prevent us from choosing
$A=B$. This technique can be formalized as follows:\eatpar

\begin{itemize}

\item show that there exists a classification, representative,
and extension functions for the problem $B$;

\item show that there exists a reduction from $B$ to $B$
satisfying representative equivalence.

\end{itemize}

The most obvious reduction from a problem to itself is the identity.
In our case, however, identity does not satisfy the condition of
representative equivalence. As a result, we have to look for
another reduction.

Before showing the technical details of the reductions used,
we point out an important feature of this technique. Since
the condition of representative equivalence tells that $B$
is \nucC-hard if $A$ is \C-hard, using $A=B$ we prove
that $B$ is \nucC-hard whenever $B$ is \C-hard. This result
holds even if a precise complexity characterization of $B$
is not known. For example, if we only know that $B$ is in
\S{2}, but do not have any hardness result, we can still
conclude that $B$ is \nucnp-hard if it is \np-hard, it
is \nucconp-hard if it is \conp-hard, it is \nucS{2}-hard
if it is \S{2}-hard, etc.

In order to simplify the following proofs, we denote with
$\Pi(X)$ the set of all distinct clauses of length 3 on a
given alphabet $X = \{x_1, \ldots, x_n\}$. Since the theory
$T$ is in 3CNF by assumption, we have that $T \subseteq \Pi(V)$,
where $V$ is the set of variables appearing in $T$.

%%%%%%%%%%%%% existence %%%%%%%%%%%%%%
\subsection{Existence of Solutions}

In order to define a reduction from the problem of
existence of solutions to itself, we first consider
the function $f$ from abduction instances to
abduction instances defined as follows:

\begin{eqnarray*}
f( \l H,M,T \r ) 
&=&
\l H',M',T' \r
\\
&&
\mbox{where:}
\\
&&
\begin{array}{rcl}
H' &=&
H \cup C \cup D \\
M' &=&
M \cup \{ c_i ~|~ \gamma_i \in T \} \cup \{ d_i ~|~ \gamma_i \not\in T \} \\
T' &=&
\{ \neg c_i \vee \neg d_i ~|~ \gamma_i \in \Pi(H \cup X)\}
\cup \{ c_i \rightarrow \gamma_i ~|~ \gamma_i \in \Pi(H \cup X) \}
\end{array}
\end{eqnarray*}

In these formulae, $X$ denotes the alphabet of $T$, while
$C$ and $D$ are sets of new variables in one-to-one correspondence
with the clauses in $\Pi(H \cup X)$. Note that, by definition,
$T$ is a subset of $\Pi(H \cup X)$. The following lemma
relates the solutions of $\l H,M,T \r$ with the solutions
of $\l H',M',T' \r$.

\begin{lemma}
\label{basic}

Let $f$ be the function defined above. For any $H$, $M$, $T$,
it holds:\eatpar

\[
SOL(f(\l H,M,T \r)) =
\{ S \cup \{ c_i ~|~ \gamma_i \in T \} \cup
\{ d_i ~|~ \gamma_i \not\in T \}
~|~ S \in SOL(\l H,M,T \r) \}
\]

\end{lemma}

\proof We divide the proof in three parts. In the first
part, we prove that any solution of $f(\l H,M,T \r)$
contains exactly the literals $c_i$ and $d_i$ that are in $M'$.
In the second part, we prove that, if $S'$ is a solution
of $f(\l H,M,T \r)$, then $S' \backslash (C \cup D)$ is
a solution of $\l H,M,T \r$; the third part is the proof
of the converse.

\begin{enumerate}

\item We prove that
$S' \cap (C \cup D) =
\{ c_i ~|~ \gamma_i \in T \} \cup \{ d_i ~|~ \gamma_i \not\in T \}$.
Let $R= \{ c_i ~|~ \gamma_i \in T \} \cup \{ d_i ~|~ \gamma_i \not\in T \}$.
Since $R \subseteq M'$, we have that $S' \cup T' \models R$.
If $c_i \in R$, then $S' \cup T' \models c_i$.
Since $T'$ does not contain any positive occurrence of $c_i$,
the theory $S' \cup T'$ can imply $c_i$ only if $c_i \in S'$.
The same holds for any $d_i \in R$. This proves that
$S' \cap (C \cup D) \supseteq R$. Since $R$ contains
either $c_i$ or $d_i$ for any $i$, the same holds for $S'$.
No other variable in $C \cup D$ can be in $S'$, otherwise
$S'$ would be inconsistent with $T'$, which contains
the clauses $\neg c_i \vee \neg d_i$.

\item Let $S'$ be an element of $SOL(\l H',M',T' \r)$. We prove that $S = S'
\backslash (C \cup D) \in SOL(\l H,M,T \r)$. The point proved above shows that,
for each $i$, $S'$ contains either $c_i$ or $d_i$, depending on whether
$\gamma_i \in T$. As a result:\eatpar

\begin{eqnarray*}
S' \cup T'
&\equiv&
S \cup \{ c_i ~|~ \gamma_i \in T \} \cup \{ d_i ~|~ \gamma_i \not\in T \}
\cup \{ \neg c_i \vee \neg d_i \} \cup \{ c_i \rightarrow \gamma_i \}
\\
&\equiv&
S \cup \{ c_i ~|~ \gamma_i \in T \} \cup \{ d_i ~|~ \gamma_i \not\in T \}
\cup T
\end{eqnarray*}

As a result, $S \cup T$ is consistent because the above formula is. Moreover,
since the above formula implies $M$, and each variable in $C \cup D$ appears
only once, it also holds $S \cup T \models M$. As a result, $S$ is a solution
of $\l H,M,T \r$.

\item Let $S \in SOL(\l H,M,T \r)$, and let
$S' = S \cup \{ c_i ~|~ \gamma_i \in T \} \cup
\{ d_i ~|~ \gamma_i \not\in T \}$.
Since $S' \cup T'$ is equivalent to
$S \cup T \cup \{ c_i ~|~ \gamma_i \in T \} \cup
\{ d_i ~|~ \gamma_i \not\in T \}$, then $S'$ 
is a solution of $\l H',M',T' \r$.

\end{enumerate}

The claim is thus proved.~\qed

This lemma shows that any abduction instance can be
converted into another one in which the set $H$ and
the theory $T$ only depends on the number of variables
of the original instance. This reduction can be used
to build a reduction satisfying the condition of
representative equivalence.

\begin{lemma}
\label{add-assumptions}

Let $c$ be a positive integer number, and let $g_c$
be the following function:\eatpar

\[
g_c(\l H,M,T \r) = \l H \cup \{ h_{|H|+1}, \ldots , h_c \}, M,
T \cup \{ x_{r+1} \vee \neg x_{r+1}, \ldots , x_c \vee \neg x_c \} \r
\]

\noindent where $r = |Var(T) \backslash H|$. It holds

\[
SOL(g_c(\l H,M,T \r)) =
\{ S \cup H' ~|~ S \in SOL(\l H,M,T \r) \mbox{ and }
H' \subseteq \{ h_{|H|+1}, \ldots , h_c \} \}
\]

\end{lemma}

\proof The instance $g_c(\l H,M,T \r)$ only differs
from $\l H,M,T \r$ because of the new assumptions
$h_{|H|+1}, \ldots , h_c$, which are not even mentioned
in $T$, and new tautological clauses to $T$. Therefore,
any explanation of $\l H,M,T \r$ is also an explanation
of $g_c(\l H,M,T \r)$. The only difference between these
two problems is that assumptions in $h_{|H|+1}, \ldots , h_c$
can be freely added to any explanations.~\qed

We now define the classification, representative, and extension functions for
the basic problems of abduction. First, the classification function is given by
the maximum between the number of variables in $H$ and the number of variables
in $T$ but not in $H$:\eatpar

\[
Class(\l H,M,T \r) = \max( |H|, |Var(T) \backslash H|)
\]

The representative instance of the class $c$ is given by an instance with $c$
possible assumptions, $c$ other variables, and $T$ composed by all possible
clauses of three literals over these variables:\eatpar

\[
Repr(c) = \l \{h_1, \ldots, h_c\}, \emptyset,
	\Pi(\{h_1, \ldots, h_c\} \cup \{x_1, \ldots, x_c\}) \r
\]

The extension function is also easy to give. For example, we may add to $T$ a
set of tautologies with new variables.\eatpar

\[
Ext(\l H,M,T \r,m) =
\l H, M, T \cup \{x_{r+1} \vee \neg x_{r+1}, \ldots, x_m \vee \neg x_m\} \r
\mbox{ where } r=|Var(T) \backslash H|
\]

These three functions are valid classification, representative,
and extension functions for the problem of existence of explanation;
they are also valid for the problems of relevance and necessity.

\

We are now able to show a reduction satisfying the condition
of representative equivalence. Let $i$ be the reduction
defined as follows.

\[
i(\l H,M,T \r) = f( g_{Class(\l H,M,T \r)} ( \l H,M,T \r) )
\]

The following theorem is a consequence of the fact that
$i$ satisfies the condition of representative equivalence.

\begin{theorem}

The problem of establishing the existence of solution of an abductive
problem is \nucS{2}-hard.

\end{theorem}

\proof
By the above two lemmas, $i(\l H,M,T \r)$ has solutions
if and only if $\l H,M,T \r$ has solution. Therefore,
$i$ is a valid reduction from the problem of solution
existence to itself. The fixed part of $i(\l H,M,T \r)$
only depends on the class of the instance $\l H,M,T \r$. As a
result, this reduction satisfies the condition of
representative equivalence. Since the problem of
existence of solutions is \S{2}-hard \cite{eite-gott-95-a},
it is also \nucS{2}-hard.~\qed

%%%%%%%%%%%%% veri-noor %%%%%%%%%%%%%%
\subsection{Verification}

We consider the problem of verifying whether a set of
assumptions is a possible explanation, still in the
case of no ordering. An instance of the problem is
composed of a triple $\l H,M,T \r$ and a specific subset
$H_a \subseteq H$ we want to check being an explanation.
Formally, this problem amounts to checking whether
$H_a \cup T$ is consistent and $H_a \cup T \models M$.
The varying part is composed of $H_a$ and $M$. Formally,
an instance of the verification problem is a 4-tuple
$\l H,H_a,M,T \r$, where $H_a \subseteq H$.

The first step of the proof is that of finding the
three functions (classification, representative,
and extension). The functions of the last proof
only require minor changes to be used now.\eatpar

\begin{eqnarray*}
Class(\l H,H_a,M,T \r) &=& \max( |H|, \var(T) \backslash H ) \\
Repr(c) &=& \l \{ h_1, \ldots, h_c \}, \emptyset, \emptyset,
	    \Pi( \{ h_1, \ldots, h_c \} \cup \{ x_1, \ldots, x_c \} \r \\
Exte(\l H,H_a,M,T \r) &=& 
	\l H,H_a,M,T \cup \{ x_{r+1} \vee \neg x_{r+1}, \ldots ,
                            x_c \vee \neg x_c \r \} \r \\
&& \mbox{where } r=| \var(T) \backslash H |
\end{eqnarray*}

We define two functions $f'$ and $g_c'$ to be similar to
the functions $f$ and $g_c$ of the last section, except
for the addition of a candidate explanation $H_a$.\eatpar

\begin{eqnarray*}
f'(\l H,H_a,M,T \r)
&=&
\l H',H_a \cup \{ c_i ~|~ \gamma_i \in T \} \cup \{ d_i ~|~ \gamma_i \not\in T \},
M',T' \r \\
&&
\mbox{ where } \l H',M',T' \r = f(\l H,M,T \r)
\\
g_c'(\l H,H_a,M,T \r)
&=&
\l H',H_a,M',T' \r
\\
&&
\mbox{ where } \l H',M',T' \r = g_c(\l H,M,T \r)
\end{eqnarray*}

These functions can be composed to generate a function
that satisfies representative equivalence. This way,
we prove the nucomp-hardness of the problem of
verification.

\begin{theorem}

The problem of verification with no ordering is \nucDp-complete.

\end{theorem}

\proof
By Lemma~\ref{basic} and Lemma~\ref{add-assumptions}, $H_a \subseteq H$ is a
solution of $g_c(\l H,M,T \r)$ if and only if it is a solution of $\l H,M,T
\r$, and that $H_a \cup \{ c_i ~|~ \gamma_i \in T \} \cup \{ d_i ~|~ \gamma_i
\not\in T \}$ is a solution of $f(\l H,M,T \r)$ if and only if $H_a$ is a
solution of $\l H,M,T \r$.

As a result, both $f'$ and $g_c'$ are reductions from the problem of
verification to itself. Moreover, their composition $i'$ satisfies
representative equivalence, since the fixed part of $i'(\l H,H_a,M,T \r)$
only depends on the class of the instance $\l H,H_a,M,T \r$. We can then
conclude that the problem of verification is hard for the
compilability class that corresponds to the complexity class
it is hard for.~\qed

%%%%%%%%%%%%% rele-noor %%%%%%%%%%%%%%
\subsection{Relevance, Dispensability, and Necessity}

We make the following simplifying assumption: given an
instance of abduction $\l H,M,T \r$, where $H = \{h_1,\ldots,h_m\}$,
the problem is to decide whether the first assumption $h_1$ is
relevant/dispensable/necessary. Clearly, the complexity of these
problems is the same, as we can always rename the variables
appropriately.

\begin{theorem}

The problems of relevance and dispensability with no ordering
is \nucS{2}-hard, while necessity is \nucP{2}-hard.

\end{theorem}

\proof
By Lemma~\ref{basic} and Lemma~\ref{add-assumptions}, 
$i(\l H,M,T \r)$ is a reduction from the problem
of relevance to the problem of relevance. Indeed, for
any $H_a \subseteq H$, the set
$H_a \cup \{ c_i ~|~ \gamma_i \in T \} \cup
\{ d_i ~|~ \gamma_i \not\in T \}$ is a solution
of $f(g_c(\l H,M,T \r))$ if and only if $H_a$ is a
solution of $\l H,M,T \r$. As a result, $h_1$ is
relevant/dispensable/necessary for $\l H,M,T \r$ if and
only if it is so for $f(g_c(\l H,M,T \r))$.

The function $i$ satisfies representative equivalence, since the fixed part of
$i(\l H,M,T \r)$ only depends on the class of $\l H,M,T \r$. What is left to
prove is the existence of the three functions. We can use the same three ones
used for the problem of existence of solutions.~\qed

%%%%%%%%%%%%% noclass %%%%%%%%%%%%%%
\section{Compilability of Abduction: Preferences}

In this section, we consider the problems of verification,
relevance, and necessity when the ordering used is either
$\leq$ or $\subseteq$. These orderings have in common the
fact that the instance of an abduction problem is simply
a triple $\l H,M,T \r$, whereas the orderings of the next
section employee classes of priority or weights that are
part of the instances. The problem of existence is the same  
as with no ordering, as these orderings are well founded.

%%%%%%%%%%%%% meth-nocl %%%%%%%%%%%%%%
\subsection{Some General Results}

We give some general results about the problem of abduction in the case in
which an ordering on explanation is given. In order to keep results as general
as possible, we consider an arbitrary ordering $\preceq$ satisfying
the following natural conditions.

\begin{description}

\item[Meaningful.] The ordering $\preceq$ is meaningful if, for any variable
$h$ and any pair of sets $H'$ and $H''$ such that $h \not\in H' \cup H''$
it holds:\eatpar

\[
H' \cup \{h\} \preceq H'' \cup \{h\} ~~ \mbox{ iff } ~~ H' \preceq H''
\]

Intuitively, a meaningful ordering compares two explanations
$H'$ and $H''$ only on the variables they differ.

\item[Irredundant] The ordering $\preceq$ is irredundant if, for any pair
of sets $H'$ and $H''$ it holds:\eatpar

\[
H' \subset H'' ~~ \Rightarrow ~~ H' \prec H''
\]

Irredundancy formalizes the natural assumption that
hypotheses that are not necessary should be removed.

\end{description}

We determine the compilability of abduction with preference
in the same way we did in the case of no ordering: we show
that the function $i$ is a polynomial reduction from the
problems of abduction to themselves, and that it satisfies
the condition of representative equivalence.
To this aim, we need the analogous of Lemma~\ref{basic} and
Lemma~\ref{add-assumptions}.

\begin{lemma}
\label{basic-order}

If $\preceq$ is a meaningful ordering, it holds:\eatpar

\[
SOL_\preceq(f(\l H,M,T \r)) =
\{ S \cup \{ c_i ~|~ \gamma_i \in T \} \cup \{ d_i ~|~ \gamma_i \not\in T \}
~|~ S \in SOL_\preceq(\l H,M,T \r) \}
\]

\end{lemma}

\proof We use the result of Lemma~\ref{basic}. Namely, since all solutions
of $f(\l H,M,T \r)$ coincide on $C \cup D$, these variables are irrelevant
thanks to the fact that $\preceq$ is meaningful.

Formally, we have:\eatpar

\begin{eqnarray*}
\lefteqn{S \in SOL_\preceq(f(\l H,M,T \r))} \\
& \Leftrightarrow &
S \in SOL(f(\l H,M,T \r)) \mbox{ and }
\not\exists S' \in SOL(f(\l H,M,T \r)) ~.~ S' \preceq S
\\
& \Leftrightarrow &
S = S_1 \cup \{ c_i ~|~ \gamma_i \in T \} \cup \{ d_i ~|~ \gamma_i \not\in T \},
\\
&&
S_1 \in SOL(H,M,T) \mbox{ and }
\\
&&
\not\exists S'_1 \in SOL(\l H,M,T \r) \mbox{ such that }
\\
&& ~~~
S'_1 \cup \{ c_i ~|~ \gamma_i \in T \} \cup \{ d_i ~|~ \gamma_i \not\in T \}
\prec S_1 \cup \{ c_i ~|~ \gamma_i \in T \} \cup \{ d_i ~|~ \gamma_i \not\in T \}
\\
& \Leftrightarrow &
S = S_1 \cup \{ c_i ~|~ \gamma_i \in T \} \cup \{ d_i ~|~ \gamma_i \not\in T \}, ~~
S_1 \in SOL(H,M,T) \mbox{ and }
\\
&&
\not\exists S'_1 \in SOL(\l H,M,T \r) ~.~
S'_1 \prec S_1
\\
& \Leftrightarrow &
S = S_1 \cup \{ c_i ~|~ \gamma_i \in T \} \cup \{ d_i ~|~ \gamma_i \not\in T \}
\mbox{ and }
S_1 \in SOL_\preceq(\l H,M,T \r)
\end{eqnarray*}

This proves the claim.~\qed

We can also prove the analogous of Lemma~\ref{add-assumptions}.

\begin{lemma}
\label{add-assumptions-order}

Let $c$ be a positive integer number and let $g_c$ be the following
function:\eatpar

\[
g_c(\l H,M,T \r) = \l H \cup \{ h_{|H|+1}, \ldots , h_c \}, M,
T \cup \{ x_{r+1} \vee \neg x_{r+1}, \ldots , x_c \vee \neg x_c \} \r
\]

\noindent where $r = |Var(T) \backslash H|$.
If $\preceq$ is an irredundant ordering, it holds:\eatpar

\[
SOL_\preceq(g_c(\l H,M,T \r)) = SOL_\preceq(\l H,M,T \r)
\]

\end{lemma}

\proof Similar to the proof of Lemma~\ref{add-assumptions},
but now the hypotheses in $\{ h_{|H|+1}, \ldots , h_c \}$
are all irrelevant; therefore, they are not part of any
minimal explanation.~\qed

These lemmas can be used to prove incompilability of abduction
when an irredundant and meaningful ordering is used.

%%%%%%%%%%%%% veri-nocl %%%%%%%%%%%%%%
\subsection{Verification}

We consider the problem of verifying whether a set of
assumptions is a minimal explanation according to the
orderings $\leq$ and $\subseteq$. More generally, we
prove the following theorem for any meaningful and
irredundant ordering.

\begin{theorem}
\label{veri-nocl}

If $\preceq$ is a meaningful and irredundant ordering, verifying whether a
set of assumptions is a minimal explanation is \nucC-hard for any class \C\
for which the problem is \C-hard.

\end{theorem}

\proof The same classification, representative, and extension
functions used for the case of no ordering can be used for
this case as well.

Let now consider the functions $f'$ and $g_c'$. From
Lemma~\ref{basic-order} and Lemma~\ref{add-assumptions-order}
it follows that they are reductions from the problem of
verification to itself. Moreover, their composition $i'$
satisfies representative equivalence.~\qed

%%%%%%%%%%%%% rele-nocl %%%%%%%%%%%%%%
\subsection{Relevance, Dispensability, and Necessity}

We make the following simplifying assumption: given an
instance of abduction $\l H,M,T \r$, where
$H = \{h_1,\ldots,h_m\}$, the problem is to decide whether
the first assumption $h_1$ is relevant/dispensable/necessary.
There is no loss of generality in making this assumption.
as we can always rename the variables appropriately.

\begin{theorem}
\label{rele-nocl}

If $\preceq$ is a meaningful and irredundant ordering, then
the problems of relevance/dispensability/necessity are
\nucC-hard for any class \C\  of the polynomial hierarchy
for which they are \C-hard.

\end{theorem}

\proof
From Lemma~\ref{basic-order} and Lemma~\ref{add-assumptions-order},
it follows that the reduction $i$ is a reduction from the problems of
relevance/dispensability/necessity to themselves, if $\preceq$ is meaningful
and irredundant, and it also satisfies representative equivalence.~\qed

Since $\subseteq$ and $\leq$ are meaningful irredundant orderings,
their complexity implies their compilability characterization.

\begin{corollary}

Relevance and dispensability using $\subseteq$ are \nucS{2}-hard,
while using $\leq$ they are \nucDlog{3}-hard. Necessity is
\nucP{2}-hard and \nucDlog{3}-hard, using $\subseteq$ and $\leq$,
respectively.

\end{corollary}

%%%%%%%%%%%%% prio %%%%%%%%%%%%%%
\section{Compilability of Abduction: Prioritization and Penalization}

We consider the cases in which the ordering over the explanations is defined in
terms of a prioritization. The instances of the problem are different from
those of the previous section, since $H$ is replaced by a partition of
assumptions $\l H_1,\ldots,H_m \r$.

In the cases of $\leq$-prioritization and $\subseteq$-prioritization,
the induced ordering $\preceq$ is meaningful and irredundant. However,
the results on meaningful irredundant ordering cannot be directly
applied because, in Theorem~\ref{veri-nocl} and Theorem~\ref{rele-nocl},
we assumed that the instances have the form $\l H,M,T \r$, while now
they have the form $\l \l H_1,\ldots,H_m \r,M,T \r$.
Therefore, we have to find new classification, representative,
and extension functions.

We first consider the problem of verification, and prove 
its nucomp-hardness. Then, we move to the problems of
relevance, dispensability, and necessity. As for the
case of $\leq$-preference and $\subseteq$-preference,
we employee a sort of normal form, in which the assumption
we check is the first one.

\subsection{Verification}

First of all, we show the classification, representative,
and extension functions for the problem of verification.
The instances of the problem include a ``candidate
explanation'' $H_a$. 

\begin{eqnarray*}
\lefteqn{Class(\l \l H_1,\ldots,H_m \r,H_a,M,T \r)}
\\
&=&
\max( m, |H_1|, \ldots, |H_m|, |\var(T) \backslash \cup H_i| )
\\
\lefteqn{Repr(c)}
\\
&=&
\l \l \{ h^1_1,\ldots,h^1_c \} , \ldots , \{ h^c_1,\ldots,h^c_c \} \r,
\emptyset, \emptyset,
\Pi( \{ h^1_1,\ldots,h^1_c \} \cup \cdots \cup \{ h^c_1,\ldots,h^c_c \} \cup
     \{ x_1,\ldots,x_c \} ) \r
\\
\lefteqn{Exte(\l \l H_1,\ldots,H_m \r,H_a,M,T \r,m)}
\\
&=&
\l \l H_1,\ldots,H_c \r, H_a, M,
T \cup \{ x_{r+1} \vee \neg x_{r+1}, \ldots, x_m \vee \neg x_m \} \r
\\
&& \mbox{ where } r = | \var(T) \backslash \cup H_i |
\end{eqnarray*}

These functions can be easily proved to be valid
classification, representative, and extension functions.
What is missing is a reduction from the problem of
verification to itself satisfying the condition of
representative equivalence.

To this extent, we use two functions $f''$ and $g_c''$
that are similar to $f$ and $g_c$, respectively. In
particular, $f(\l \l H_1,\ldots,H_m \r,H_a,M,T \r) =
\l  \l H_1',\ldots,H_m' \r, H_a', M', T' \r$, where:\eatpar

\begin{eqnarray*}
H_1' &=& H_1 \cup C \cup D \\
H_2' &=& H_2 \\
& \vdots & \\
H_m' &=& H_m \\
H_a' &=& H_a \cup \{ c_i ~|~ \gamma_i \in T \} \cup \{ d_i ~|~ \gamma_i \not\in T \} \\
M' &=& M \cup \{ c_i ~|~ \gamma_i \in T \} \cup \{ d_i ~|~ \gamma_i \not\in T \} \\
T' &=& \{ \neg c_i \vee \neg d_i ~|~ \gamma_i \in \Pi( \var(T) \cup \bigcup H_i ) \} \cup
       \{ c_i \rightarrow \gamma_i ~| ~ \gamma_i \in \Pi( \var(T) \cup \bigcup H_i ) \}
\end{eqnarray*}

Besides the partition of the assumptions, this is exactly the function used
in Lemma~\ref{basic}. As a result, we have that:\eatpar

\[
SOL(\l \l H_1',\ldots,H_m' \r, M', T' \r) =
\{ S \cup \{ c_i ~|~ \gamma_i \in T \} \cup \{ d_i ~|~ \gamma_i \not\in T \} ~|~
S \in SOL (\l \l H_1,\ldots,H_m \r, M, T \r ) \}
\]

Since $\preceq$ is a meaningful irredundant ordering, the
same property holds replacing $SOL$ with $SOL_\preceq$.
The last step is to define a function $g_c''$ similar to
$g_c$. This is done as follows.\eatpar

\begin{eqnarray*}
\lefteqn{g_c''(\l \l H_1,\ldots,H_m \r, H_a, M, T \r) =} \\
&&
\l \l H_1 \cup \{ h^1_{|H_1|+1}, \ldots, h^1_c \} ,
      \ldots,
      H_m \cup \{ h^m_{|H_m|+1}, \ldots, h^m_c \} ,
      \ldots,
      \{ h^c_1,\ldots,h^c_c \} \r,
   H_a, \\
&& M,
   T \cup \{ x_{r+1} \vee \neg x_{r+1}, \ldots, x_c \vee \neg x_c \} \r \\
&& \mbox{ where } r = | \var(T) \backslash \bigcup H_i |
\end{eqnarray*}
   
In words, each $H_i$ is extended with new assumptions to make
it contain exactly $c$ assumptions. Some new classes of assumptions
$H_i$ are added, in such a way the resulting instance contains
exactly $c$ classes of assumptions. Finally, $T$ is extended
with tautologies over new variables, in such a way the variables of the
new theory that are not assumptions are exactly $c$.

The resulting instance is defined in such a way all its
relevant numbers (number of classes of assumptions, number of
assumptions in each class, number of other variables in the theory)
coincide. The analogous of Lemma~\ref{add-assumptions-order} holds:
the solutions of $\l \l H_1,\ldots,H_m \r, H_a, M, T \r$ and the
solutions of $g_c''(\l \l H_1,\ldots,H_m \r, H_a, M, T \r)$ coincide.
This is due to the fact that $g_c''$ only introduces new variables
that are irrelevant to the minimal solutions.

\begin{theorem}

The problem of verification for any prioritization based on a meaningful and
irredundant ordering is \nucC-hard for any class \C\  for which it is \C-hard.

\end{theorem}

\proof
The composition of $i''$ of $f''$ and $g_c''$ is a reduction
satisfying representative equivalence.

\[
i''(\l \l H_1,\ldots,H_m \r, H_a, M, T \r) =
f( g''_{Class(\l \l H_1,\ldots,H_m \r, H_a, M, T \r)} 
(\l \l H_1,\ldots,H_m \r, H_a, M, T \r) )
\]

The fact that is a reduction from the problem of verification to itself easily
follows from the fact that both $f''$ and $g_c''$ are. Moreover, the result of
$f( g''_{Class(\l \l H_1,\ldots,H_m \r, H_a, M, T \r)} (\l \l H_1,\ldots,H_m
\r, H_a, M, T \r) )$ is an instance in which the number of classes of
assumption, of variables in each class, and the number of other variables, all
coincide with the class of the original instance. The function $f''$ produces
an instance in which everything but $M$ and $H_a$ depends only on these
numbers. As a result , the function $i$ produces an instance in which
everything but $M$ and $H_a$ depends on the class of the original instance
only. As a result, this function $i''$ is a reduction from the problem of
verification to itself, satisfying representative equivalence, which implies
the incompilability of the problem.~\qed

\subsection{Relevance and Necessity}

We restrict the problems to the case the assumption we
want to check for relevance/dispensability/necessity
is the first variable of $H_1$. The problems have the
same complexity of the general ones (in which the assumption
can be an arbitrary one.) This, however, cannot be proved
with a simple renaming of the variables, as we did for
the case of preference.

\begin{theorem}
\label{first-of-first}

Let $\preceq$ be a meaningful and irredundant ordering.
It holds:\eatpar

\begin{eqnarray*}
& h^i_j \mbox{ is relevant/necessary for }
\l \l H_1,\ldots,H_m \r, M, T \r & \\
& \mbox{ iff } & \\
& t \mbox{ is relevant/necessary for } \\
&
\l \l \{t, s\}, H_1,\ldots,H_m \r, M \cup \{u,v\}, T \cup
\{ h^i_j \rightarrow u, t \rightarrow v,
s \rightarrow u, s \rightarrow v \} \r &
\end{eqnarray*}

\end{theorem}

\proof We first give an informal sketch of the proof.
The set of solutions (with no ordering) of
the first and the second instances only differ because
the explanations for the second instances must contain
either $s$ or both $h^i_j$ and $t$.

The explanations of the second instances are first compared
on the assumptions in $H_1,\ldots,H_m$, and then on
$\{s,t\}$. Therefore, the ordering for the second instance
is a refinement of the ordering of the first one. Namely,
a minimal solution of the second instances is either a
minimal solution of the first one plus $s$, or a minimal
solution of the first one plus $t$. However, the latter is
a solution only if it contains $h^i_j$. Therefore, the
presence of a solution containing $h^i_j$ in the first
instance is equivalent to the presence of a solution for
the second instance containing $t$.

The formal proof is as follows. Let $\l H,M,T \r$ be the
first instance and $\l H',M',T \r$ be the second one.

\begin{enumerate}

\item $S' \in SOL(H',M',T')$ implies
$S' \backslash \{s,t\} \in SOL(H,M,T)$.

This can be proved as follows. First, since $S' \cup T'$ is
consistent, it follows that $S' \cup T$ is consistent as well
(because $T \subseteq T'$), which also implies that
$(S' \backslash \{s,t\}) \cup T$ is consistent.

Let us now prove that $(S' \backslash \{s,t\}) \cup T \models M$. By
assumption, we have $S' \cup T' \models M'$. The following
chain of implications leads to the claim.

\begin{eqnarray*}
& S' \cup T' \models M \cup \{u, v\} & \\
& \Downarrow & \\
& S' \cup T' \models M & \\
& \Downarrow & \\
& S' \cup T' \cup \neg M \mbox{ is inconsistent} & \\
& \Downarrow & \\
& (S' \backslash \{s,t\}) \cup T \cup (S' \cap \{s,t\}) \cup
\{ h^i_j \rightarrow u, t \rightarrow u, t \rightarrow v, s \rightarrow v \}
\cup \neg M 
\mbox{ is inconsistent} & \\
& \mbox{ since $u$ and $v$ appears only positively, set $u=y=\true$ } & \\
& \Downarrow & \\
& (S' \backslash \{s,t\}) \cup T \cup (S' \cap \{s,t\}) \cup 
\neg M \mbox{ is inconsistent} & \\
& \mbox{ $s$ and $t$ appears (at most) once: they can be removed } & \\
& \Downarrow & \\
& (S' \backslash \{s,t\}) \cup T \cup \neg M \mbox{ is inconsistent} & \\
& \Downarrow & \\
& (S' \backslash \{s,t\}) \cup T \models M &
\end{eqnarray*}

\item $S' \in SOL_\preceq(H',M',T')$ implies
$S' \backslash \{s,t\} \in SOL_\preceq(H,M,T)$.

Proved by reductio ad absurdum. Assume that
$S' \in SOL_\preceq(H',M',T')$, but that
$S' \backslash \{s,t\} \not\in SOL_\preceq(H,M,T)$.
As proved above, $S' \backslash \{s,t\} \in SOL(H,M,T)$.
As a result, it is not minimal: there exists
another $S'' \in SOL(H,M,T)$ such that
$S''$ is better than $S' \backslash \{s,t\}$.
As proved above, $S'' \cup \{s,t\} \in SOL_\preceq(H',M',T')$.
Moreover, $S'' \cup \{s,t\}$ is better than $S'$, because
$s$ and $t$ are in the lowest class of the prioritization.

\item If $h^i_j \in S$, then $S \in SOL_\preceq(H,M,T)$ if and only if
$S \cup \{t\} \in SOL_\preceq(H',M',T')$.

By the point 1 and 2 above, if $S \cup \{t\}$ is a minimal solution of the
second instance, then $S$ is a minimal solution of the first one. We prove the
converse.

First of all, $S \cup \{t\}$ is solution of the second instance. What is left
to prove is its minimality. This is also easy: removing $t$ leads to a set
of assumptions which does not explain $v$. If removing some variable from $S$
leads to another solution, then $S$ is not minimal.

\item If $h^i_j \not\in S$, then $S \in SOL_\preceq(H,M,T)$ if and only if
$S \cup \{s\} \in SOL_\preceq(H',M',T')$.

The ``if'' direction is easy. Let us assume that $S$ is a minimal solution of
the first instance. Then $S \cup \{s\}$ is a solution of the second one. Let us
prove that it is minimal. We cannot remove variables from $S$, otherwise $S$
would be not minimal. As a result, the only other possible explanations that
can be preferred are $S \cup \{t\}$ and $S \cup \emptyset$. None of them is a
solution, because they do not imply $u$.

\end{enumerate}

It is now possible to prove the claim. If there exists a minimal solution
of the first instance containing $h^i_j$, then there exists a minimal solution
of the second one containing $t$. On the other hand, if no minimal solution
contains $h^i_j$, then all corresponding minimal solutions of the second
instances contains $s$, which means that $t$ is in none of them.
Therefore, relevance and necessity of $h^i_j$ on the first
instance are equivalent to relevance and necessity, respectively,
of $t$ in the second instance.~\qed

As a result of this theorem, we can assume that relevance or dispensability are
evaluated w.r.t.\  the first variable in $H_1$. In order to prove that these
problems are not compilable, we give a classification, representative, and
extension function.\eatpar

\begin{eqnarray*}
&& Class( \l \l H_1,\ldots,H_m \r, M, T \r )
=
\max( m, |H_1|, \ldots, |H_m|, |Var(T) \backslash (H_1 \cup \cdots \cup H_m)| )
\\
&& Repr(c)
=
\l \l \{h^1_1,\ldots,h^1_c\},\ldots,\{h^c_1,\ldots,h^c_c\} \r, \emptyset, \\
&& ~~~~~~~~~~~~~
\Pi( \{h^1_1,\ldots,h^1_c\} \cup \cdots \cup \{h^c_1,\ldots,h^c_c\} \cup
\{x_1,\ldots,x_c\}) \r
\\
&&Ext( \l \l H_1,\ldots,H_m \r, M, T \r, m )
= \l \l H_1,\ldots,H_m \r, M,
T \cup \{ x_{r+1} \vee \neg x_{r+1}, \ldots, x_m \vee\neg x_m \} \r \\
&& ~~~~~~~~~~~~~
\mbox{ where } r = |Var(T) \backslash (H_1 \cup \cdots \cup H_m)|
\end{eqnarray*}

Given these three functions, all is needed is a reduction from the problem of
relevance to itself satisfying representative equivalence. The function $i''$
cannot be used only because the instance it deals with contains the set of
assumptions $H_a$. However, removing this part of the instance both from its
argument and its result, we obtain a reduction with the right
properties. We can thus conclude that the problems of relevance,
dispensability, and necessity are incompilable.

\begin{theorem}

Let $\preceq$ be a meaningful and irredundant ordering. The problems of
relevance, dispensability, and necessity for the problem of prioritized
abduction are \nucC-hard for any class \C\  of the polynomial hierarchy for
which these problems are \C-hard.

\end{theorem}

As a result, we easily obtain the compilability properties of the problem
of prioritized abduction using the orderings $\subseteq$ and $\leq$.

\begin{theorem}

Relevance and dispensability are \nucS{3}-hard if $\subseteq$ is used,
and \nucD{3}-hard if $\leq$ is used instead.

\end{theorem}

The compilability of relevance and dispensability in the case of penalization
is an easy consequence of the last theorem, as relevance with $\leq$
(prioritized) can be directly translated (using a nucomp reduction) to
relevance with penalization.

\begin{corollary}

Relevance and dispensability are \nucD{3}-hard, in the case of penalization.

\end{corollary}

\let\subsectionnewpage=\newpage

%%%%%%%%%%%%% horn %%%%%%%%%%%%%%
\section{The Horn Case}

The Horn case can be dealt with using the same technique
of the general case. Since, however, only Horn clauses
are allowed, each time we use $\Pi(H \cup X)$, which
contains all clauses of three literals over $H \cup X$,
we have to replace it with the $\Pi_H(H \cup X)$ that
contains all Horn clauses of three literals over the
set $H \cup X$. The reductions we used employ clauses
$\neg c_i \vee \neg d_i$ and $\neg c_i \vee \gamma_i$,
which are Horn if $\gamma_i$ is Horn. The reduction
used in Theorem~\ref{first-of-first} also involves Horn
clauses only. Therefore, all results holding for the
general case hold for the Horn case as well. Namely,
all problems about Horn clauses are \nucC-hard for the
same classes \C\  they are \C-hard.
An important feature of reduction from the same problem
is that it allows proving nucomp-hardness result even
for a restriction of the problem, provided that these
reduction do not transform an instance into a non-valid
one (\eg, unless an Horn instance is mapped into a
non-Horn one.)

The even more restricted case of definite Horn clauses,
however, cannot be dealt with in the same manner. Indeed,
the clauses $\neg c_i \vee \neg d_i$, are not definite.
Some problems, however, becomes polynomial, in this case.
Namely, all problems in the case of no order are polynomial,
as well as necessity for $\subseteq$-preference. We only
show that a reduction for the case of $\leq$-preference.
As before, the problem is that of checking whether $h_1$
is in an explanation of minimal size of $\l H,M,T \r$.
Since $h_1$ is part of $H$, we regard $\l H,M,T \r$ as
being the instance of the problem. The classification,
representative, and extension functions are as usual
(tautologies are definite Horn clauses.)

The reduction we use is based on the following function
$f$, where $n = |H|$.

\begin{eqnarray*}
f(\l H,M,N \r) &=&
\l H',M',T' \r
\\
&& \mbox{where} \\
&&
\begin{array}{rcl}
H' &=&
H \cup \{ c^j_i ~|~ \gamma_i \in \Pi_H(X \cup H), ~ 1 \leq j \leq n+1 \}
\\
M' &=&
M \cup \{ c^j_i ~|~ \gamma_i \in T, ~ 1 \leq j \leq n+1 \}
\\
T' &=&
\{ \gamma_i \vee \bigvee \{ \neg c^j_i ~|~  1 \leq j \leq n+1 \} ~|~
\gamma_i \in \Pi_H(X \cup H) \}
\end{array}
\end{eqnarray*}

The idea is simply that of replicating each variable $c^j_i$
for $n+1$ times. This way, if $S \subseteq H$, then a clause
$\gamma_i$ holds in $S \cup T$ only if $S$ contains all
clauses $c^j_i$.

The reduction is based on the following two facts:

\begin{enumerate}

\item definite Horn clauses are always consistent with
sets of positive literals;

\item checking the existence of explanations is polynomial.

\end{enumerate}

Therefore, the instance $\l H,M,T \r$ can be solved by
first checking whether it has explanations. If it has,
we can reduce it to $\l H',M',T' \r$. Being both $T$ and
$T'$ definite Horn theories, consistency is not an issue.
In other words, $S \subseteq H$ is an explanation of the
first instance if and only if $S \cup T \models M$, and
the same for the second instance.

\begin{lemma}

For any $C' \subseteq C$ such that $|C'| \leq n$,
it holds that $S$ is an explanation of $\l H,M,T \r$ if
and only if $S \cup \{c^j_i ~|~ \gamma_i \in T \} \cup C'$
is an explanation of $f(\l H,M,T \r)$.

\end{lemma}

\proof The definition of explanation for definite
Horn clauses is: $S \subseteq H$ is an explanation
if and only if $S \cup T \models M$. Consistency
is not relevant, as any definite Horn theory is
consistent with any set of positive literals.

Let us first assume that $S$ is an explanation
of $\l H,M,T \r$, that is, $S \cup T \models M$.
Since $\{c^j_i ~|~ \gamma_i \in T \} \cup T'$
implies $\{c^j_i ~|~ \gamma_i \in T \} \cup T$,
we conclude that $S \cup \{c^j_i ~|~ \gamma_i \in T \} \cup T$
implies $M \cup \{c^j_i ~|~ \gamma_i \in T \}$.
The set $S \cup \{c^j_i ~|~ \gamma_i \in T \}$
is therefore an explanation because the latter
set is indeed $M'$. The set $C'$ is not relevant
to this part of the proof.

Let us now assume that
$S' = S \cup \{c^j_i ~|~ \gamma_i \in T \} \cup C'$
is an explanation of $f(\l H,M,T \r)$. Since $|C'| \leq n$,
then $S'$ does not contain all $c^j_i$ for any $i$.
Therefore, all clauses that are not in $T$ contains
at least an unassigned $c^j_i$ in $S' \cup T'$.
Therefore, these clauses are cannot be used to derive
a single literal in $M'$. As a result,
$S \cup \{c^j_i ~|~ \gamma_i \in T \} \cup T' \models M'$.
This is equivalent to $S \cup T \models M$, that is,
$S$ is an explanation of $\l H,M,T \r$.~\qed

This lemma can be used to relate the minimal
explanations of the two instances.

\begin{lemma}

If $\l H,M,T \r$ has explanations, then
$S$ is one of its minimal explanation if
and only if
$S \cup \{c^j_i ~|~ \gamma_i \in T \}$ is
a minimal explanation of $f(\l H,M,T \r)$.

\end{lemma}

\proof The lemma above implies that $S$
is an explanation if and only if 
$S \cup \{c^j_i ~|~ \gamma_i \in T \}$ is
an explanation, as this is the case of
$C'=\emptyset$. Let us now prove that
the minimality of these two explanations
coincide.

Let us first assume that $S$ is a minimal
explanation. We prove that
$S'=S \cup \{c^j_i ~|~ \gamma_i \in T \}$
is a minimal explanation of $\l H',M',T' \r$.
By the lemma above, $S'$ is an explanation;
we have therefore only left to prove that
it is of minimal size. Assume that $S''$
is another explanation of $\l H',M',T' \r$.
By construction, $S''$ contains
$\{c^j_i ~|~ \gamma_i \in T \}$. Therefore,
$S''$ can be smaller than $S'$ only if
$S'' \backslash \{c^j_i ~|~ \gamma_i \in T \}$
is smaller than $S' \backslash \{c^j_i ~|~ \gamma_i \in T \}$.
Since the latter coincide with $S' \cap H$,
whose size is bounded by $n$, we have that
$|S'' \backslash \{c^j_i ~|~ \gamma_i \in T \}| \leq n$.
Therefore, $S''$ can be written as
$S''=S''' \cup \{c^j_i ~|~ \gamma_i \in T \} \cup C'$
with $|S''' \cup C'| < |S|$. The latter inequality
implies $|C'| < n$: by the lemma above, $S'''$
would be an explanation of $\l H,M,T \r$.
Since $|S''' \cup C'| < |S|$, then $|S'''| < |S|$,
that is, $S$ is not be minimal.

Let us now assume that
$S'=S \cup \{c^j_i ~|~ \gamma_i \in T \}$
is a minimal explanation of $\l H',M',T' \r$,
and prove that $S$ is a minimal explanation
of $\l H,M,T \r$. Assume, indeed, that
$S''$ is a smaller explanation of $\l H,M,T \r$.
By the lemma above, $S'' \cup \{c^j_i ~|~ \gamma_i \in T \}$
would then be an explanation of $\l H',M',T' \r$
smaller than $S'$.~\qed

The reduction can be defined as for the Horn
case, by taking into account the fact that
the original instance $\l H,M,T \r$ may not
have any explanation. Such a reduction ratifies
the condition of representative equivalence,
thus proving that problems about $\leq$-preference
are \nucC-hard whenever they are \C-hard. Similar
reductions can be defined for the other orderings.

%%%%%%%%%%%%% concl %%%%%%%%%%%%%%
\section{Conclusions}

In this paper, we have shown that logic-based
abduction cannot be simplified by preprocessing
the theory $T$ and the hypotheses $H$. In
particular, this result holds for various kinds
of explanation orderings, and also for the
Horn restriction. These results have been proved
using the technique of representative equivalence
\cite{libe-01-jacm}; since reductions are from
a problem to itself, they prove that a problem is
``compilability-hard'' for any class for which
it is hard. In other words, we did not prove that
a problem is hard for some class, but rather that
it complexity decreases thanks to preprocessing.
Using these ``self-reductions'' allows for proving
such a result even if the complexity of the problem
is not known. For example, we prove that a
preprocessing step does not simplify the problem
of finding a minimal explanation for any ordering
that is both meaningful and irredundant. The
complexity of this problem is not known for
all such orderings; moreover, it depends on the
ordering itself.

The technique we used to prove that ``preprocessing
does not simplify abduction'', being based on
complexity classes at last, should however not
be considered as implying that preprocessing is
not useful for speeding up solving of abduction
problems. Indeed, as for any result based on the
theory of \np-completeness, this conclusion
only holds as a worst-case result. In other words,
it does not tell that no instance can ever by
made simpler by preprocessing, but simply that any
preprocessing procedure necessarily has some hard
instances that are not simplified. In a sense,
our result is more positive than it appears, as it
tells that a worst-case exponential on-line
algorithm is reasonable, given than no
worst-case polynomial one exists.

Compilability results based on hardness and reductions
have consequences similar to complexity results based
on the theory of \np-completeness: they tell that,
since no worst-case polynomial algorithm can solve
the problem, alternative directions have to be considered.
Approximation is one example: the preprocessing phase
may result in some data structure that allows a
better (or faster) approximation of the best
abductive explanations. Another possible direction
is that of incomplete compilation, in which the
preprocessing phase produces a result that is only
useful in some cases, but not always. Another common
solution to hard-to-compile problems is that of
generating a worst-case exponential preprocessing
result. This approach is especially useful if part
of the result can be used, as we can then try to
generate it and use only the part we can store. All
these alternative approaches, however, only make
sense when the impossibility of preprocessing the
problem into a polynomial problem has been proved.
This is the practical impact of our hardness results.

Finally, compilability has been proved to be related
to expressibility of logical formalisms, that is, their
ability of representing information in little space
\cite{cado-etal-00}. Logical-based abduction formalisms
could then be characterized by the set of abductive
problems they are able to express. Compilation  classes
(and not complexity ones) have been proved useful to
this aim.

\let\sectionnewpage=\newpage

\bibliographystyle{alpha}

\end{document}